\documentclass[11pt]{amsart}

\usepackage{graphics}
\usepackage{graphicx}
\usepackage{epsfig}

\usepackage{amssymb}
\newcounter{bla}

\begin{document}

%\begin{frontmatter}

%% Title, authors and addresses

%% use the tnoteref command within \title for footnotes;
%% use the tnotetext command for the associated footnote;
%% use the fnref command within \author or \address for footnotes;
%% use the fntext command for the associated footnote;
%% use the corref command within \author for corresponding author footnotes;
%% use the cortext command for the associated footnote;
%% use the ead command for the email address,
%% and the form \ead[url] for the home page:
%%
%% \title{Title\tnoteref{label1}}
%% \tnotetext[label1]{}
%% \author{Name\corref{cor1}\fnref{label2}}
%% \ead{email address}
%% \ead[url]{home page}
%% \fntext[label2]{}
%% \cortext[cor1]{}
%% \address{Address\fnref{label3}}
%% \fntext[label3]{}

\title[Understanding Deep Neural Networks Using TDA]{Understanding Deep Neural Networks Using Topological Data Analysis}

%% use optional labels to link authors explicitly to addresses:
%% \author[label1,label2]{<author name>}
%% \address[label1]{<address>}
%% \address[label2]{<address>}

\author{Daniel Goldfarb}

\address{Cornell University, Ithaca, NY \& Air Force Research Laboratory, 711th Human Performance Wing Airman Systems Directorate, Dayton, OH}

\date{\today}
\email{dg393@cornell.edu}

%\end{frontmatter}
\maketitle

%%%%%%%%%%%%%%%%%%%

\noindent
\textbf{Abstract.}
Deep neural networks (DNN) are black box algorithms. They are trained using a gradient descent back propagation technique which trains weights in each layer for the sole goal of minimizing training error. Hence, the resulting weights cannot be directly explained. Using Topological Data Analysis (TDA) we can get an insight on how the neural network is thinking, specifically by analyzing the activation values of validation images as they pass through each layer. \\

\noindent
\textbf{Introduction.}
I will begin this paper by describing the deep neural network architecture developed by the Visual Geometry Group at Oxford. Then I will give a brief overview of TDA and the tools for using the methods in real applications. Lastly I will give a description of my research and how we can use Ayasdi's platform to analyze activation values of validation images to get some insight on how the neural network is thinking. I conclude by giving some ideas and next steps for how to use this insight to improve the overall accuracy of deep neural networks. \\

\noindent
\textbf{VGG16 Deep Neural Network.}
VGG16 \cite{1} is a deep neural network architecture developed by a team of data scientists at Oxford. It is trained on the ImageNet \cite{2} database with the goal to identify the main objects that appear in the image out of 1,000 possible labels. It performed with 70.5\% accuracy when trying to label the most prominent object in the images and 90.0\% accuracy when providing a label for one of the top 5 most prominent objects in the 2014 ImageNet competition \cite{3}. Neural networks are black box algorithms so we will be using TDA to explain how the neural network is thinking.

VGG16 consists of convolutional, pooling, and fully connected layers (16 in total). Since it is a pretrained model, all of the weights have been set in place to give the best accuracy possible. To get an insight as to what image characteristics these weights learned, we will analyze the activation values of two of the layers. The activation values are the outputs of sending each feature vector through a layer in the neural network. In a convolutional layer, a high activation value implies that the weight matrix found a characteristic in the image that it was looking for. This characteristic could be anything from a solid edge to a dark spot in the image. \\

\noindent
\textbf{Description of Data Used.}
We have narrowed the ImageNet database down to 500 images for each of 5 breeds of cats and 5 breeds of dogs, giving us 5,000 total images to work with. As features we will be using the 100,000+ activation values from the thirteenth convolutional layer and the third fully connected layer. So each image is represented as a point in 100,000+ dimensional space. It is not feasible to manually analyze such a high dimensional data set. \\

\noindent
\textbf{Topological Data Analysis Overview and Tools.}
Currently the two major areas of TDA are persistent homology and the Mapper algorithm. In my 2014 paper, I give an overview of persistent homology and a use case on National Hockey League statistics \cite{4}. The Mapper algorithm is used to summarize information about datasets using filter functions defined on the data and partial clustering \cite{5}. Ayasdi, Inc. has a commercial implementation of the Mapper algorithm which I use for my research. \\

\noindent
\textbf{Description of Research.}
We uploaded this dataset to Ayasdi's TDA platform, enabling us to understand the shape of the data, as illustrated in Figure 1. \\

\begin{center}
\includegraphics[scale=0.17]{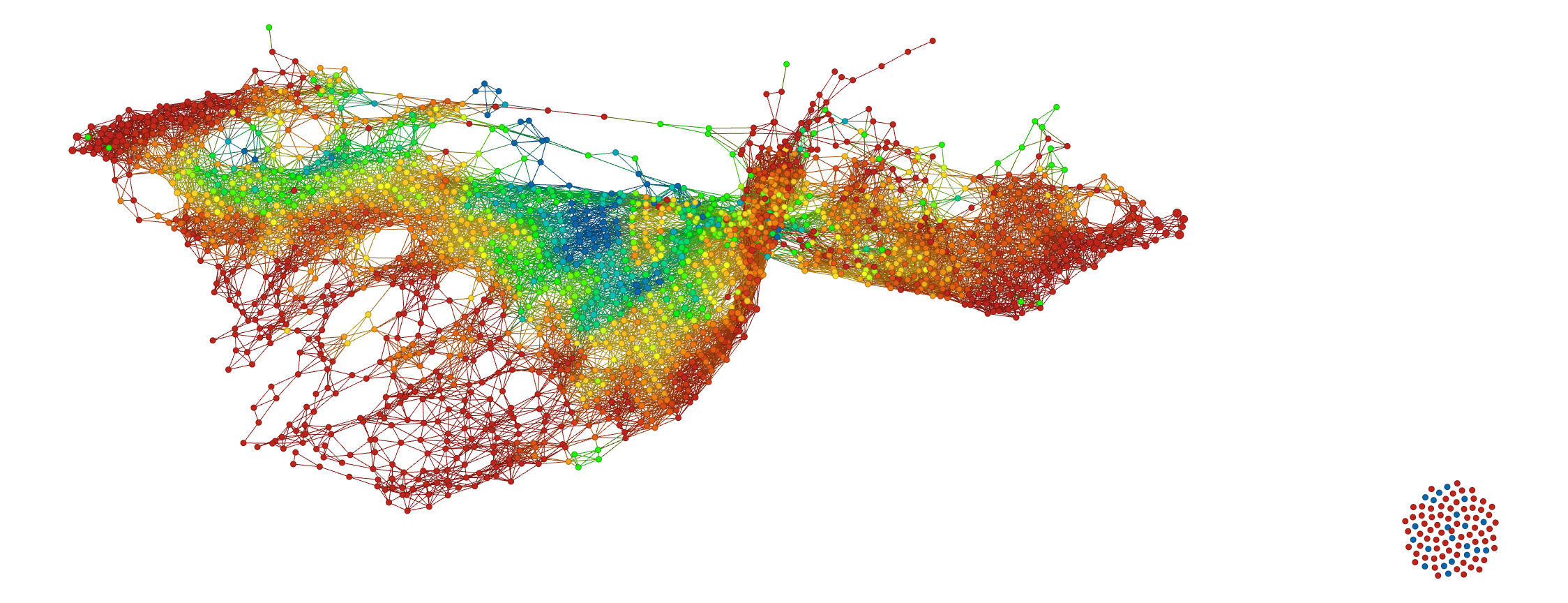}
Figure 1: The Topological Model of the subset of the ImageNet database
\end{center}

\vspace{0.7cm}

Ayasdi allows the customization of several parameters to produce a 2-dimensional model that represents the ``shape'' of the data. The 5,000 input images are represented by data points and consolidated into nodes based on a dimensionality reduction function called a lens and on a metric which determines the distance between two points. Any two nodes are connected by an edge if they share at least one point in common. In order to represent the data how we would like, there are also options for resolution which determines the total number of nodes in the model and gain which determines the number of total edges in the graph. Here we used the Variance Normalized Euclidean metric with PCA lenses, a resolution of 50, and a gain of 3. This model is colored by classification accuracy. Red nodes identify accurate classifications of the cats and dogs, while blue nodes identify inaccurate classifications. \\

We developed an Adaptive Dashboard visualization with JavaScript, enabling us to conveniently select and view groups of nodes in a topological model, as shown in Figure 2.

\begin{center}
\includegraphics[scale=0.27]{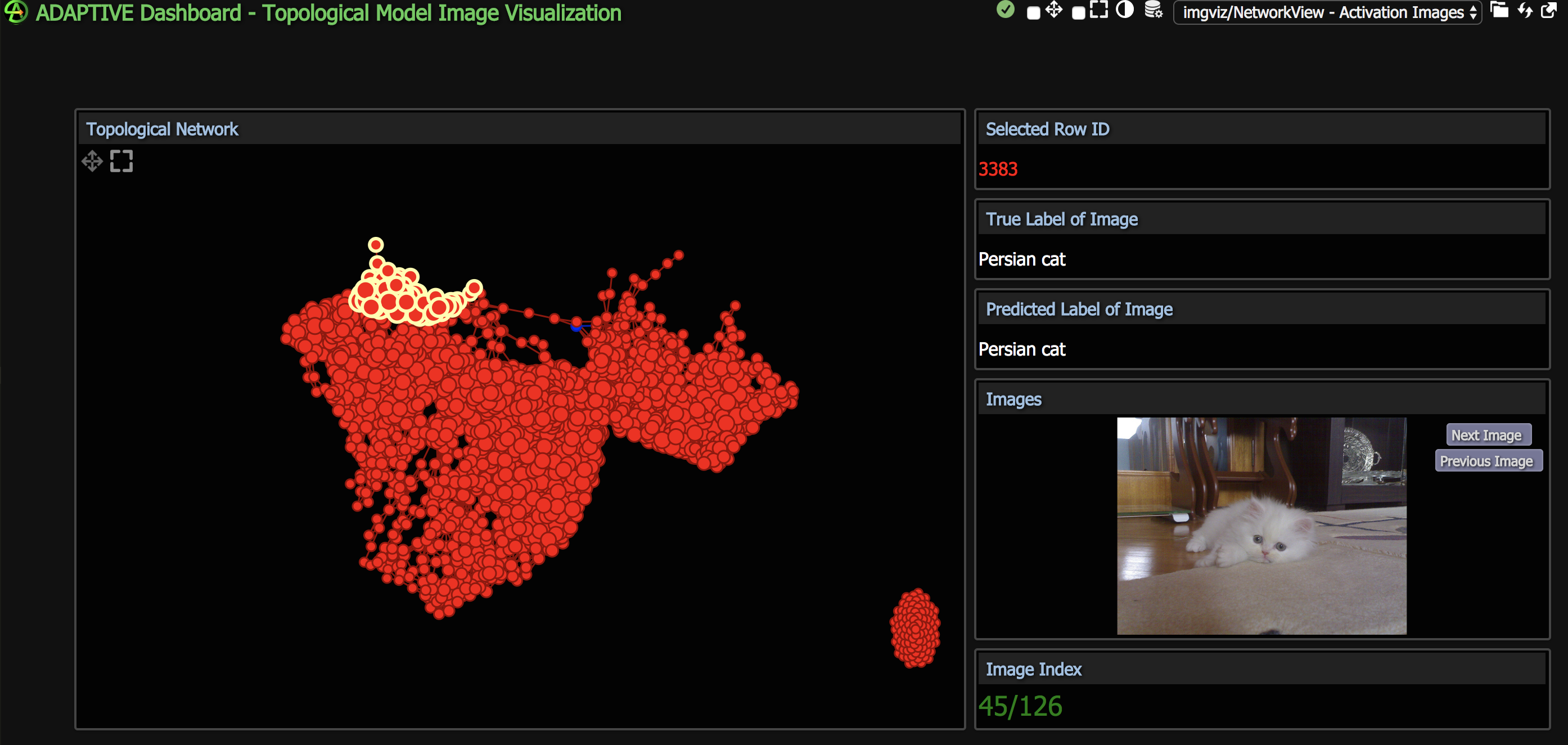}
Figure 2: Adaptive Dashboard Visualization
\end{center}

\vspace{0.7cm}

When we highlight nodes in the model on the left, they are loaded and displayed in the gauge on the right. We are then able to toggle through the images that correspond to the highlighted nodes and analyze the predictions made compared to the ground truth labels. We can now begin to understand the similarities between images that were misclassified.

One area of interest is the blue middle cluster of poor accuracy. \\

\begin{center}
\includegraphics[scale=0.5]{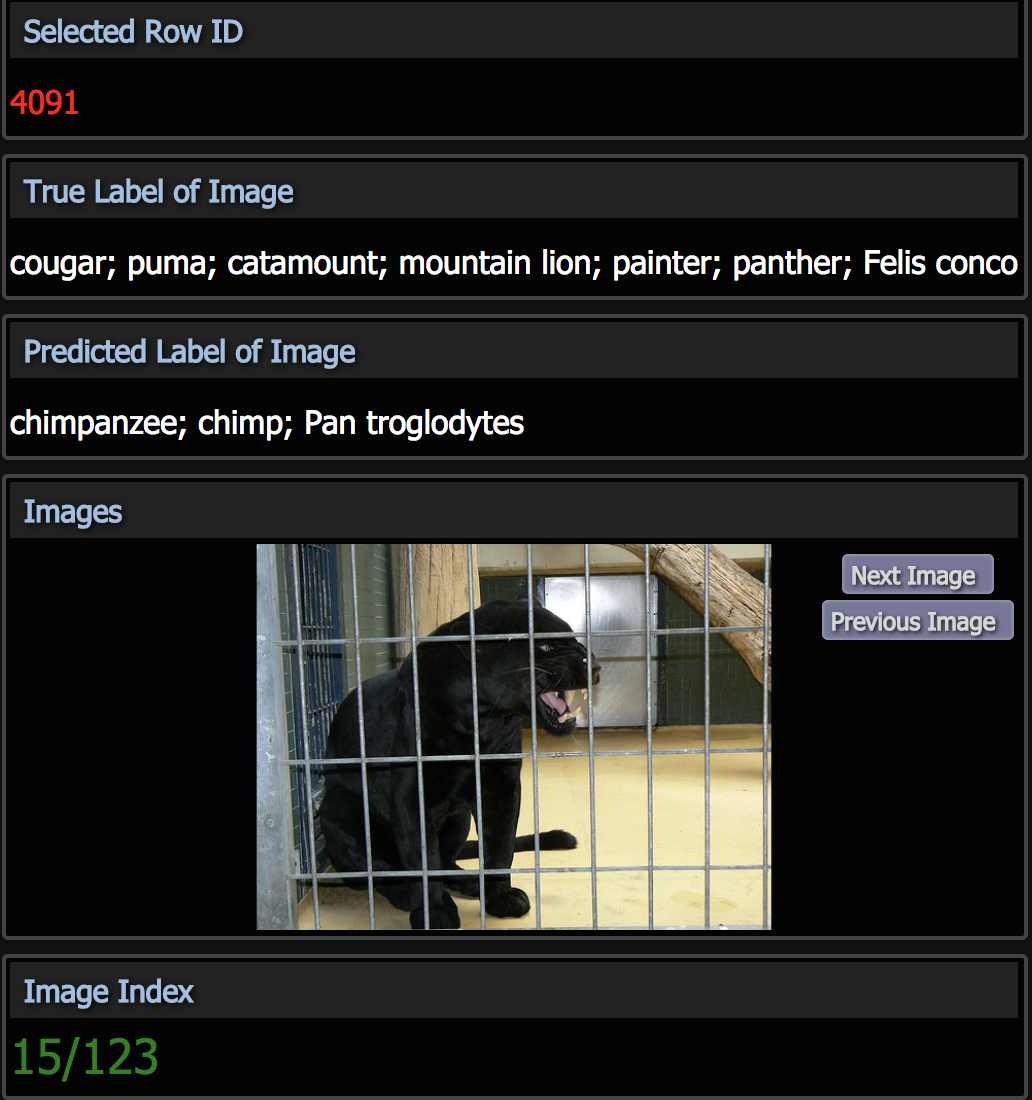} \\
Figure 3: Cougar gauge
\end{center}

As we can see from Figure 3, this cluster includes several cougars and panthers in cages. The neural network falsely labels these images as chimpanzees and other types of apes. We hypothesize that the training set included many images of chimps in zoos. So when we input an image of a cage or enclosure, the neural network assumes that we are looking at some kind of primate. We can use this knowledge of the image to give us a prior confidence on the prediction before we even input it to the neural network. For example, if the image contains a cage then we have a lower confidence in the neural network output than if the animal was in its natural habitat. A solution to this problem would be to include more images of large cats in cages in the training set so the neural network could have more experience with this specific kind of environment.

Now consider the top-left red cluster of highly accurate classifications. \\

\begin{center}
\includegraphics[scale=0.5]{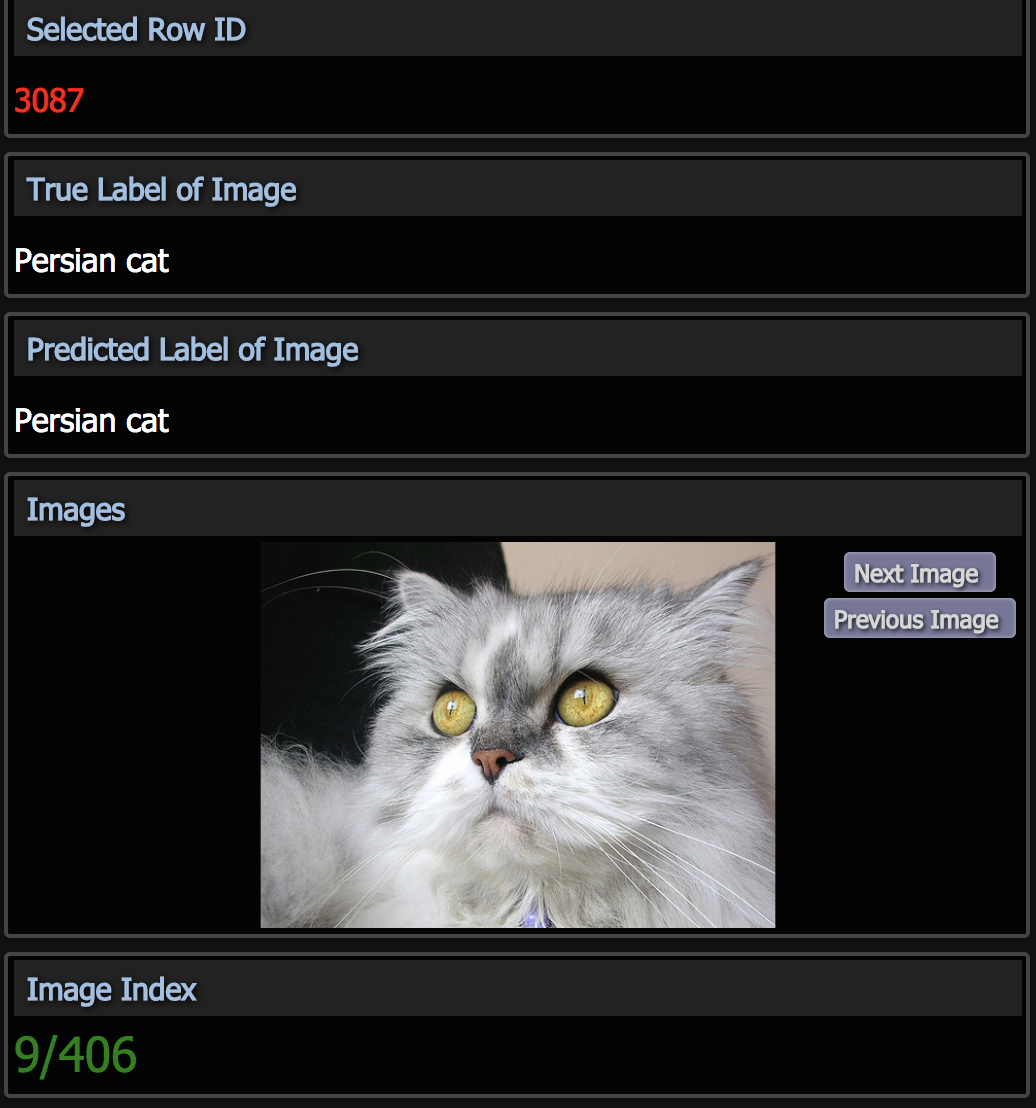} \\
Figure 4: Persian cat gauge
\end{center}

From Figure 4, we can see that this cluster includes close up images of Persian cats with plain backgrounds. In the situation where we encounter a validation image with a uniform background, we can have high confidence in VGG16 making a correct prediction. This makes sense as the neural network now has no other option than to focus on the main object in the image which is the cat itself.

Instead of making assumptions about what the neural network may be focusing on, suppose we had a way of knowing exactly which areas on the image were producing the highest activation values. We can do such analysis with the use of heat maps \cite{6}. \\ 

\begin{center}
\includegraphics[scale=0.7]{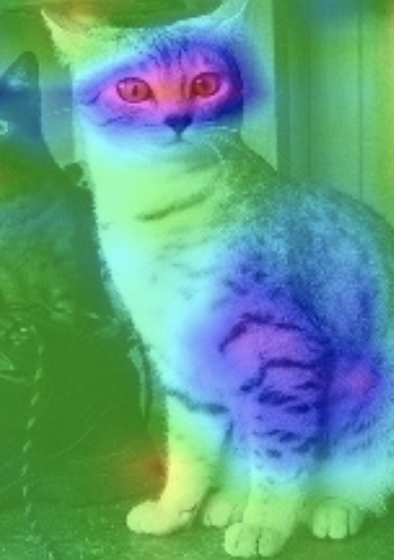} \\
Figure 5: Heat map example
\end{center}

In this heat map in Figure 5, the pink colored spots are the locations where the neural network gave relatively high activations at nodes. So we can infer that VGG16 was focusing on the cat's eyes and stripes when classifying this picture. \\

\begin{center}
\includegraphics[scale=0.37]{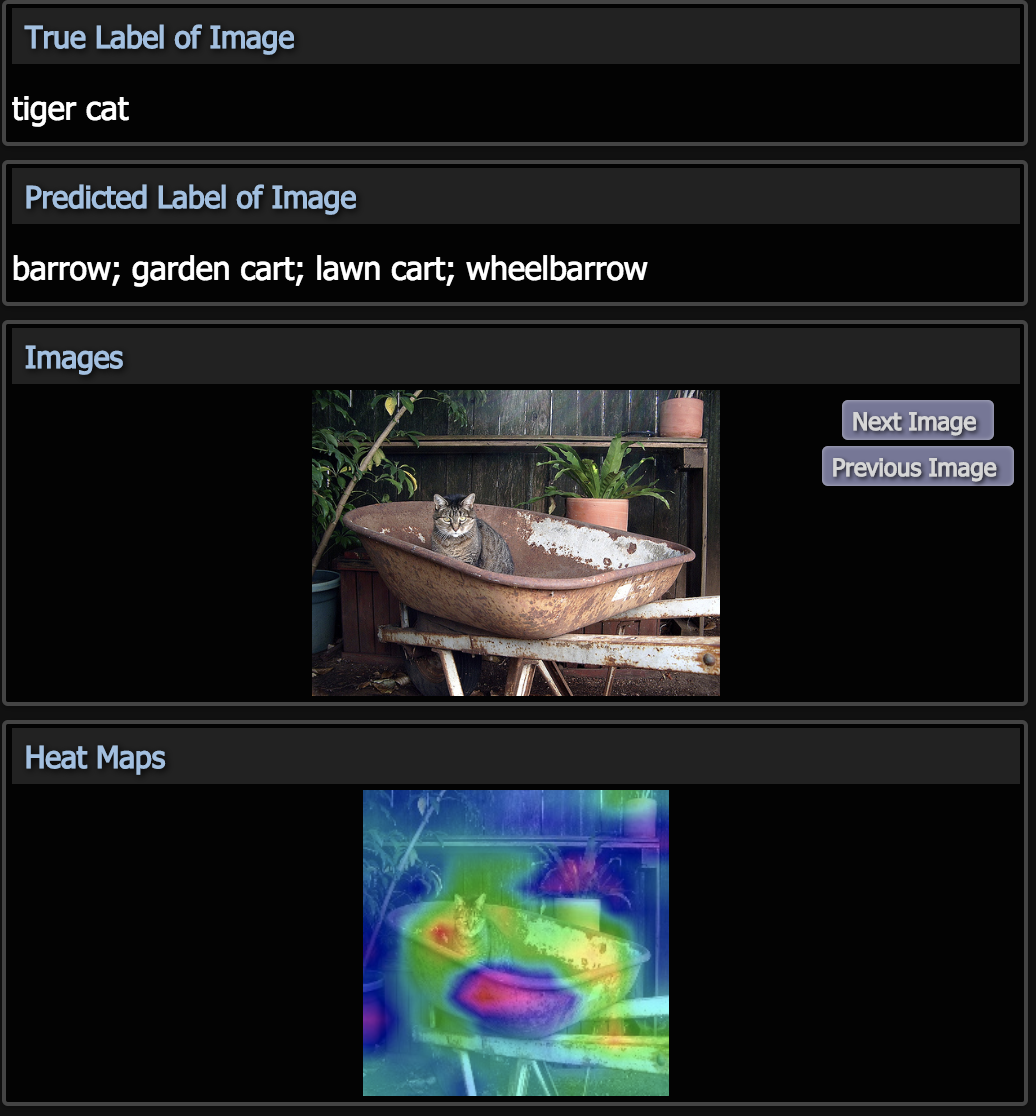} \\
Figure 6: Tiger cat heat map gauge
\end{center}

Similarly in Figure 6, we can see that the neural network was focusing on the wheelbarrow when trying to classify this image of a tiger cat. \\

\begin{center}
\includegraphics[scale=0.37]{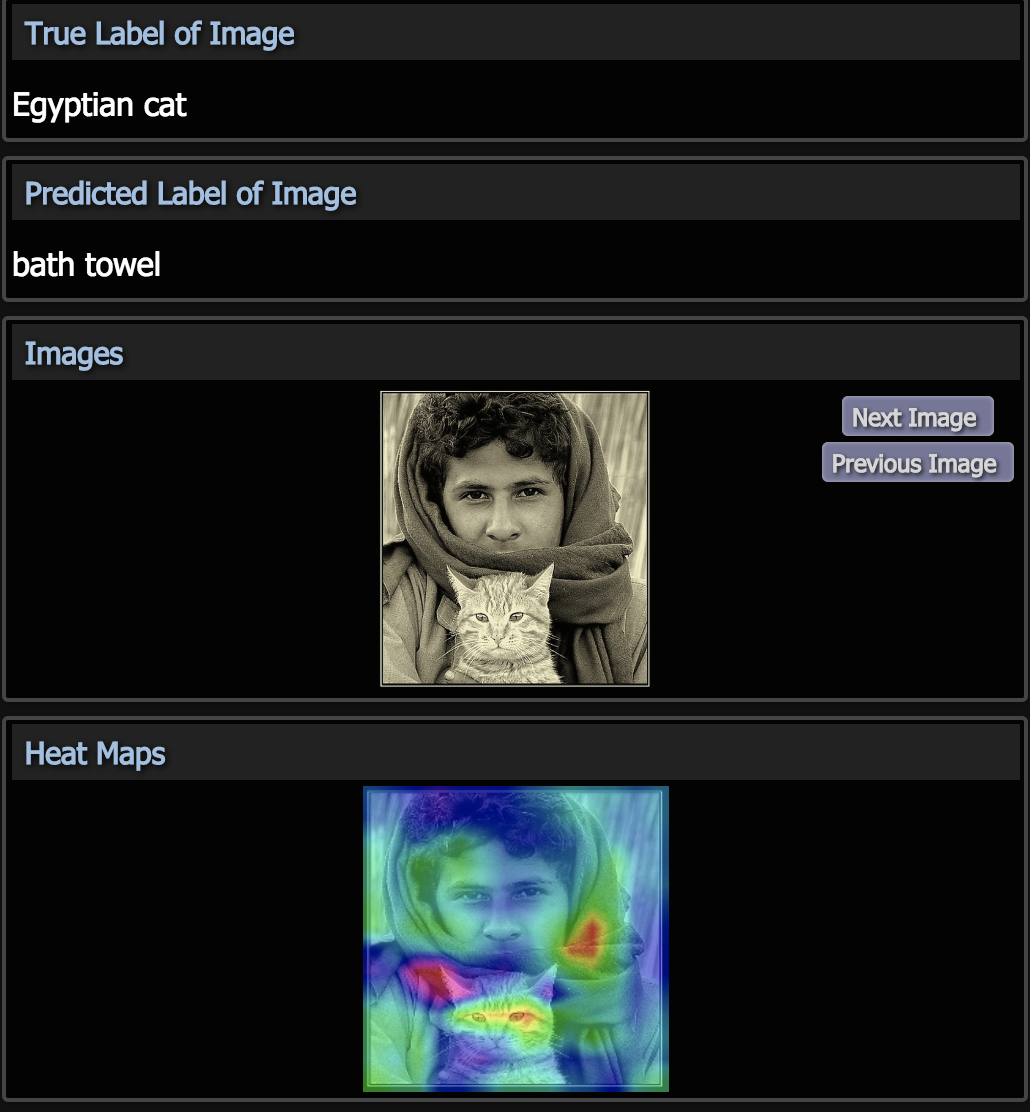} \\
Figure 7: Egyptian cat heat map gauge
\end{center}

\vspace{0.7cm}

In Figure 7 we can also see that it was focusing on the scarf when trying to classify this image of an Egyptian cat.

Instead of assuming that the neural network was distracted by other aspects of the image, we have actual proof that the highest activations were in fact focused on other objects than the cats.

\vspace{0.5cm}

\begin{center}
\includegraphics[scale=0.5]{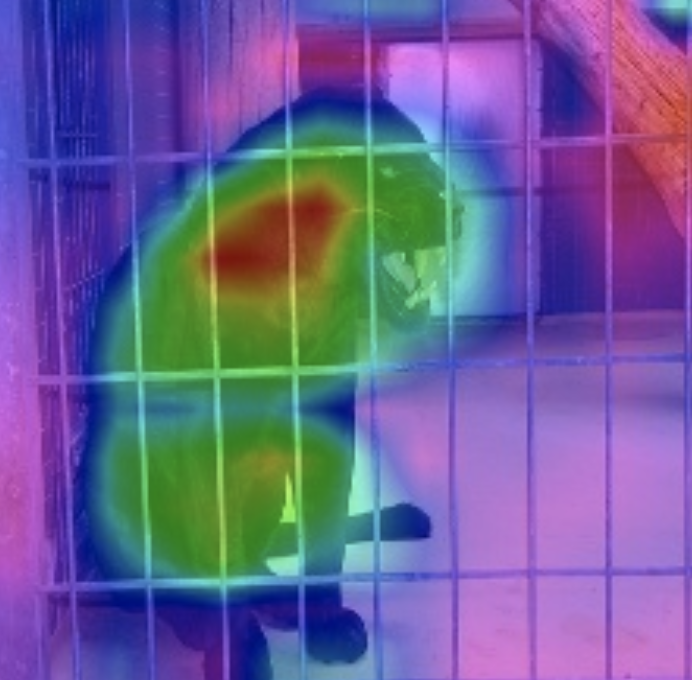} \\
Figure 8: Cougar heat map
\end{center}

\vspace{0.7cm}

In Figure 8 we now see the heat map image of the enclosed cougar. We are now confident that VGG16 was much more focused on the cage formation in the image than the cougar itself.

So far we showed how to use TDA to identify clusters of correct and incorrect classifications. When we built the model using the activation values at several layers, we noticed that the misclassifications clustered into a single dense area. It is more interesting to analyze multiple clusters for which the reason for poor classification is different for each one.

I will demonstrate this concept by showing the work we did using the pixel data of the images before the activation values had been processed. We converted 2,000 randomly picked images and extracted 5,000 pixel features. In this model the images represented a majority of the 1,000 possible ImageNet labels. Each of the test images was also given an extra one-hot feature for being "problematic". We defined a problematic label to be one that we encountered at least three times in the test set and had an overall accuracy of less than 40\%.

\vspace{0.3cm}

\begin{center}
\includegraphics[scale=0.5]{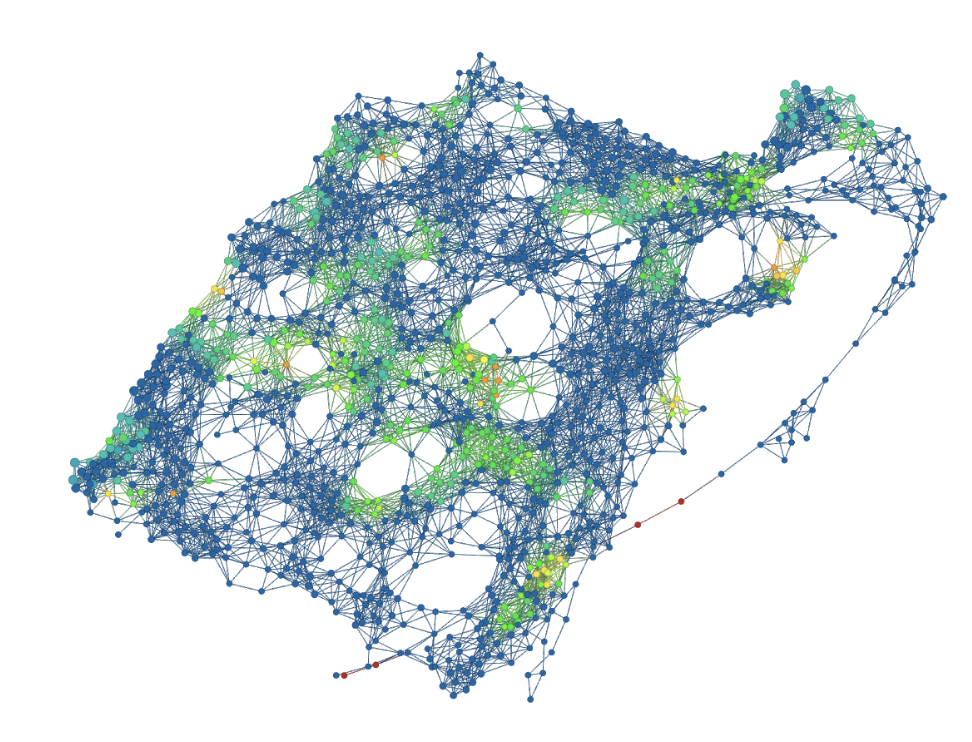} \\
Figure 9: The Topological Model using pixel features
\end{center}

\vspace{0.7cm}

The topological model in Figure 9 is colored by the density of problematic labels. The green nodes contain quite a few images that had problematic labels while blue nodes had almost none. We can see in Figure 9 that when we build the topological model, the images with problematic labels cluster into roughly 4 distinct areas. \\

\vspace{0.3cm}

\begin{center}
\includegraphics[scale=0.5]{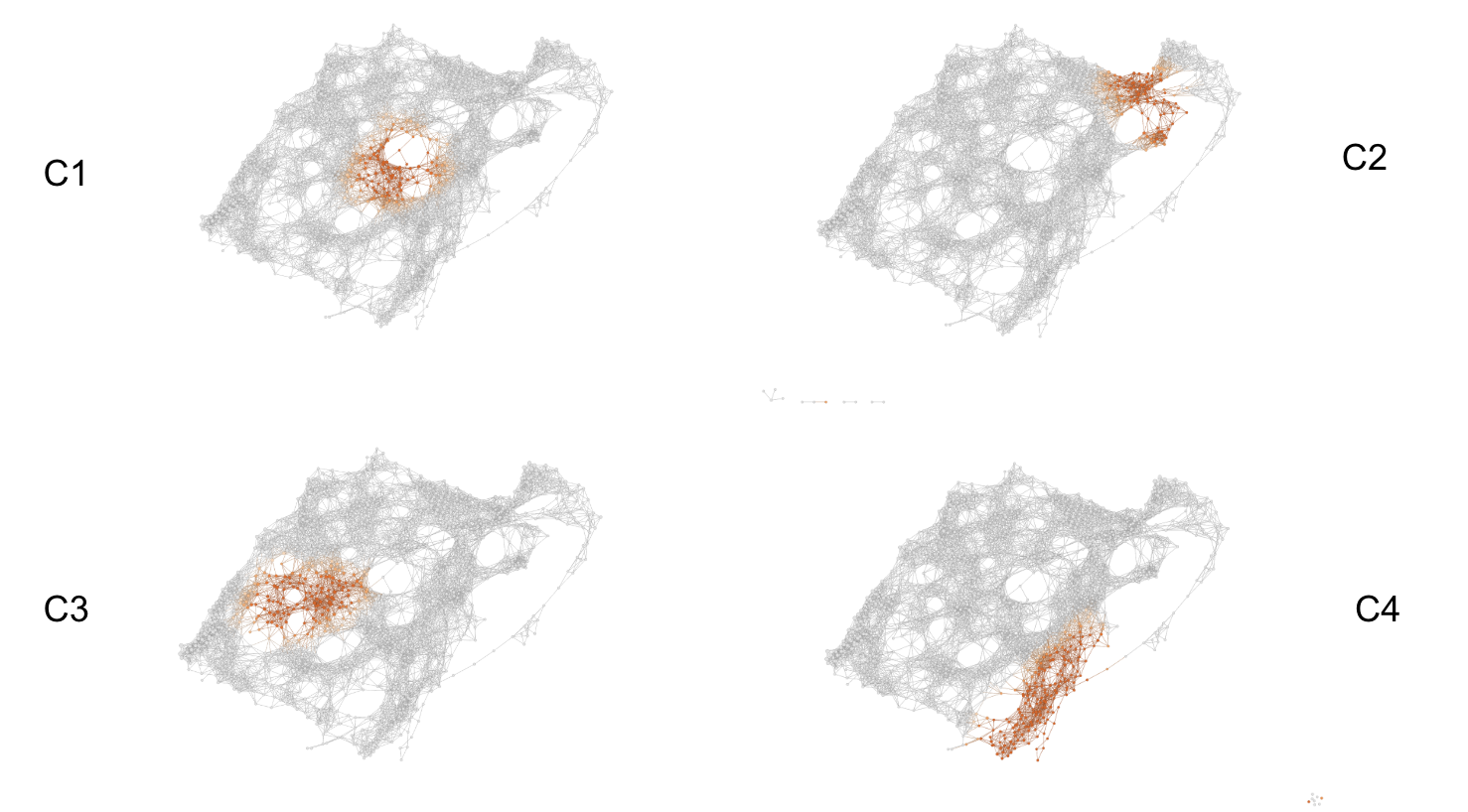} \\
Figure 10: Colorings of the four inaccurate clusters
\end{center}

\vspace{0.7cm}

Figure 10 displays four different colorings identifying exactly which of these clusters we will be looking into.

\vspace{0.3cm}

\begin{center}
\includegraphics[width=0.95\textwidth]{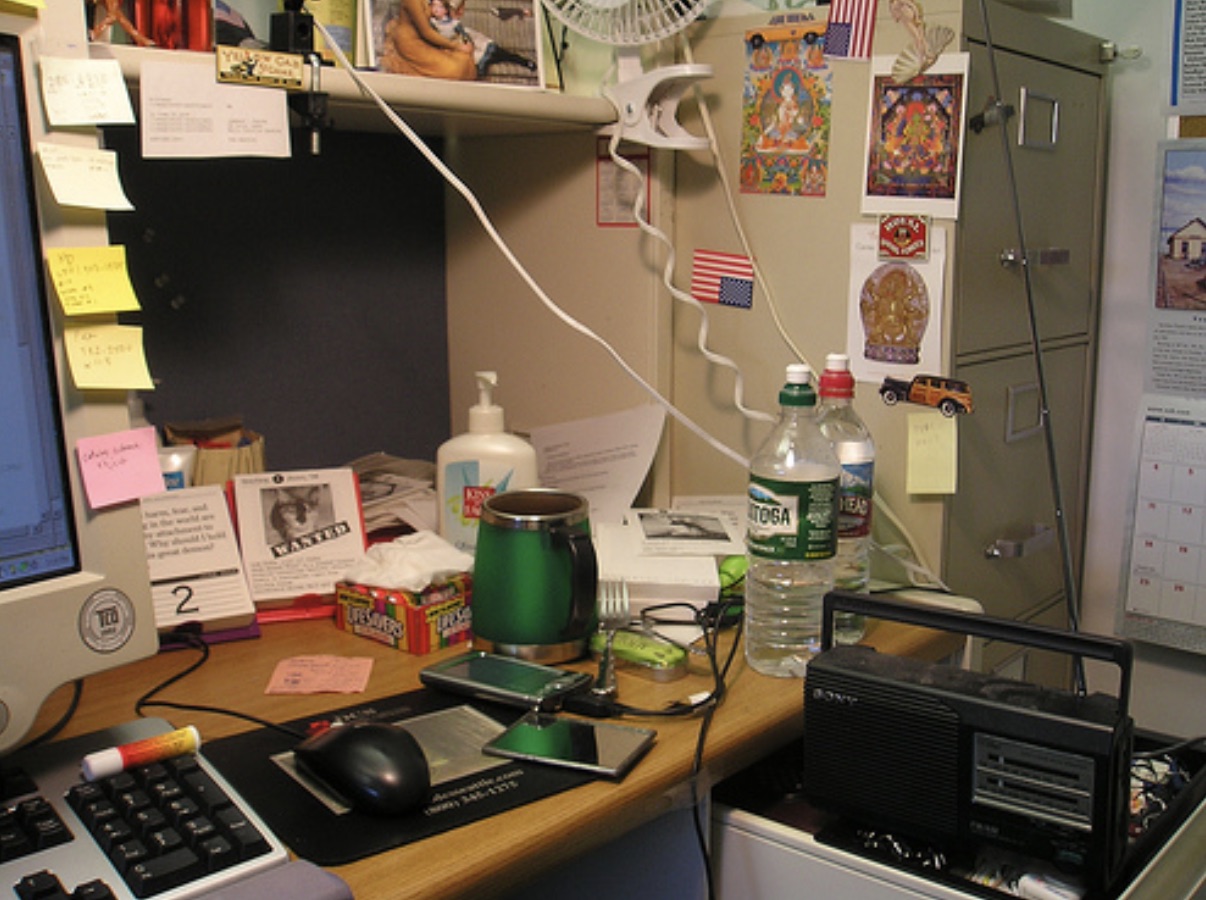} \\
Figure 11: Cluster 1 sample image
\end{center}

\vspace{0.7cm}

The category of pictures in Cluster 1 were ordinary pictures with cluttered or unusual backgrounds. In Figure 11 we have a desk but since there is so much happening on the surface and walls, VGG16 has higher confidence in the presence of the drawer than the desk as a whole.

\vspace{0.3cm}

\begin{center}
\includegraphics[width=0.95\textwidth]{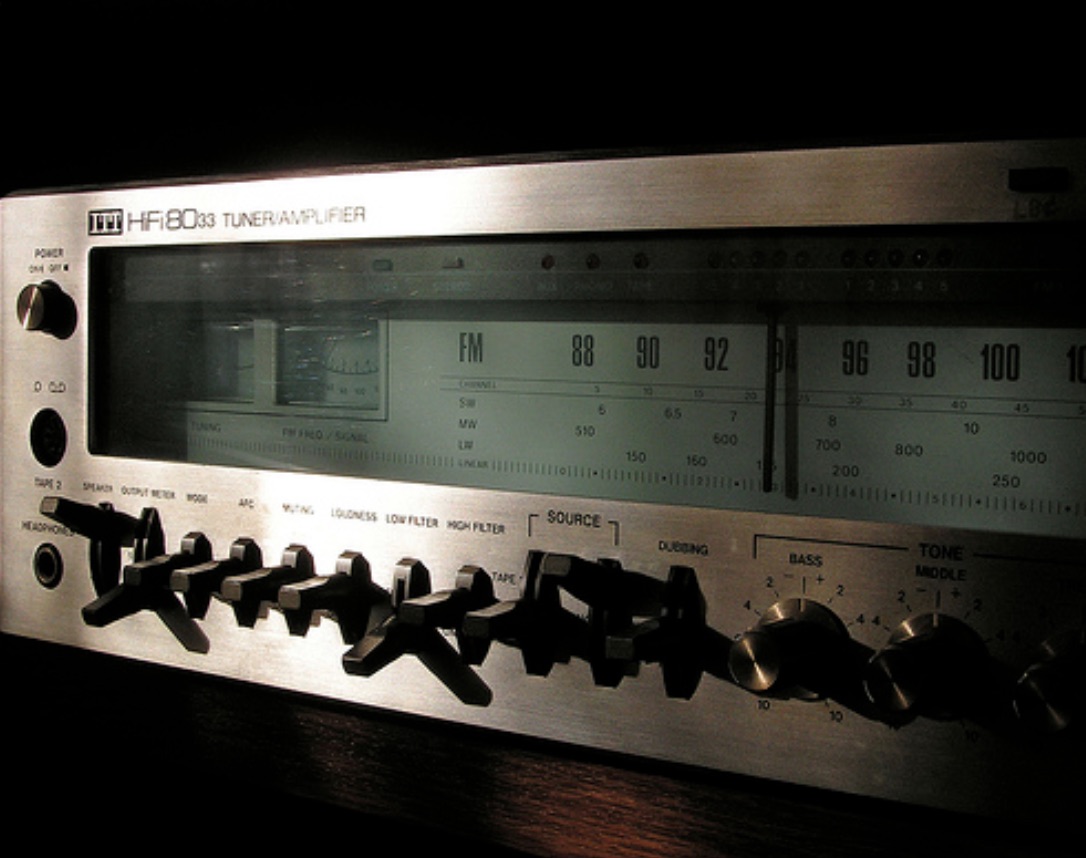} \\
Figure 12: Cluster 2 sample image
\end{center}

\vspace{0.6cm}

Cluster 2 consists of images that are shiny or have glare. These types of images could have been rare in the training set so VGG16 automatically associates glare to a small set of classes. Figure 12 shows a picture of a radio that was falsely classified as a car fender.

\vspace{0.3cm}

\begin{center}
\includegraphics[width=0.95\textwidth]{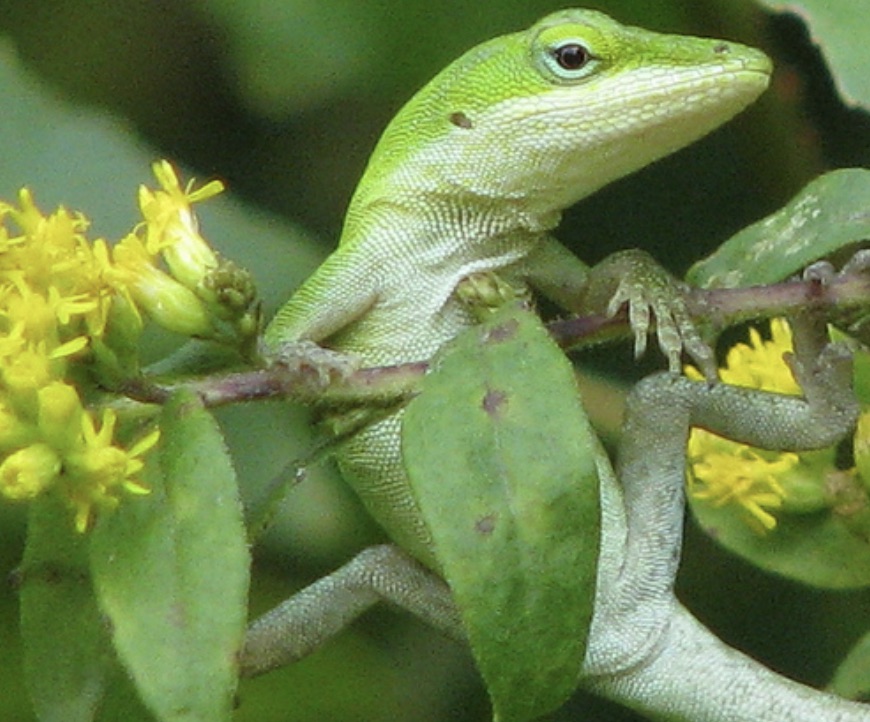} \\
Figure 13: Cluster 3 sample image
\end{center}

\vspace{0.7cm}

Cluster 3 contains several chameleons like the one in Figure 13 that have falsely been labelled as green lizards. This is likely because chameleons tend to blend in with their backgrounds similarly to how these green lizards blend in with the leafy environment. We can infer that there were few pictures of chameleons in the training set with green colored backgrounds.

\vspace{0.5cm}

\begin{center}
\includegraphics[width=0.95\textwidth]{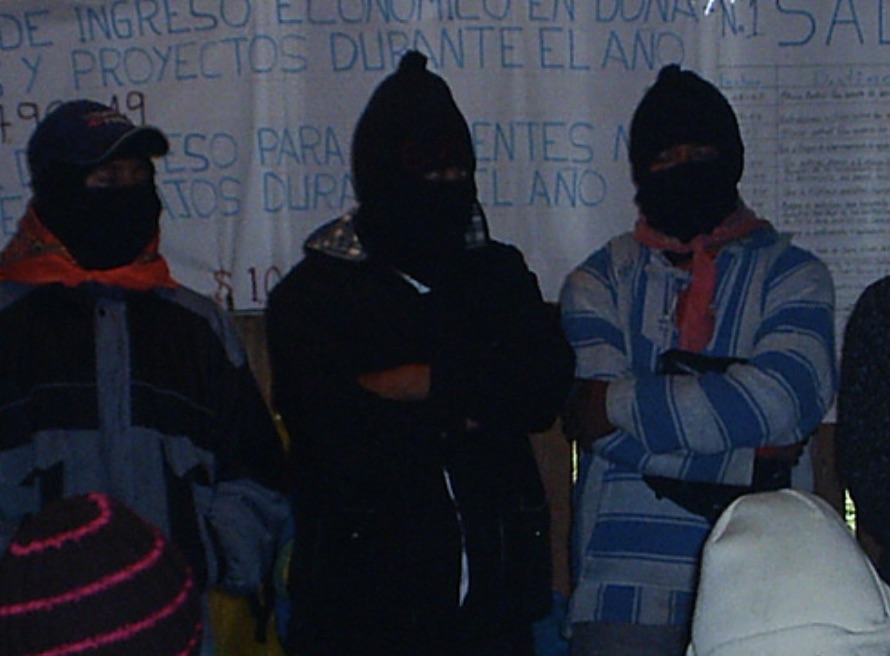} \\
Figure 14: Cluster 4 sample image
\end{center}

\vspace{0.7cm}

In Cluster 4 we run into images with shadows and very low saturations. Here VGG16 faults when its identification of certain objects relies on color. We get false identifications when there is confusion caused by shadows and lack of color.

When we are able to identify unique clusters of incorrect classifications like in this example, we can then train new learners to perform well on each of these clusters. We would know to rely on this new learner when we detect that a test image exhibits similar features as the cluster's topological group. In this case, the features were the pixel data of the images but we believe that the same methods would apply when analyzing the activation values as well. \\

\noindent
\textbf{Next Steps:}
With this new information on how the algorithm performs, we can potentially improve how we make predictions. We can find multiple clusters of images that make misclassifications for different reasons. Our next step would be to train new learners to perform well on the specific clusters that misclassify for certain reasons. This new learner would be used when a test image shares similar properties to those in the specific topological group. In the most recent example we would have 5 total learners. We would use one for each of the 4 problematic clusters and the original pretrained VGG16 for the rest of the training examples. In addition, we can add images to the training set of the neural network that exhibit properties of each of the misclassified clusters. By doing this, VGG16 itself could improve how it classifies images that are similar to those in the problematic clusters. We could also use the information about the density of misclassified images in clusters to get a better idea of the confidence of the prediction made. \\

\noindent
\textbf{Acknowledgements:}
This work has been supported by the Air Force Research Lab (AFRL). Thank you to Dr. Ryan Kramer at AFRL for providing guidance and mentoring during my work as an intern at the Summer of TDA. Thank you to Matthew Broussard for the idea and generation of the heat map images.
\vspace{1cm}

%%%%%%%%%%%%%%%%%%%

%% main text
%\section{}
%\label{}

%% The Appendices part is started with the command \appendix;
%% appendix sections are then done as normal sections
%% \appendix

%% \section{}
%% \label{}

%% References
%%
%% Following citation commands can be used in the body text:
%% Usage of \cite is as follows:
%%   \cite{key}         ==>>  [#]
%%   \cite[chap. 2]{key} ==>> [#, chap. 2]
%%

%% References with bibTeX database:

\nocite{*}
\bibliographystyle{elsarticle-num}
\bibliography{sample}

%% Authors are advised to submit their bibtex database files. They are
%% requested to list a bibtex style file in the manuscript if they do
%% not want to use elsarticle-num.bst.

%% References without bibTeX database:

% \begin{thebibliography}{00}

%% \bibitem must have the following form:
%%   \bibitem{key}...
%%

% \bibitem{}

% \end{thebibliography}

\end{document}